%% file: acl2023.tex
\pdfoutput=1

\documentclass[11pt]{article}

\usepackage[]{ACL2023}

\usepackage{times}
\usepackage{latexsym}
\usepackage{CJK}
\usepackage{enumitem}
\usepackage[T1]{fontenc}

\usepackage{microtype}

\usepackage{inconsolata}
\usepackage{times}
\usepackage{latexsym}
\usepackage{graphicx}
\usepackage{booktabs}
\usepackage{colortbl}
\usepackage{tablefootnote}
\usepackage{multirow}
\usepackage{microtype}
\usepackage{amsfonts}
\usepackage{inconsolata}
\usepackage{titlesec}
\usepackage{multirow}
\usepackage{color}
\usepackage{makecell}
\usepackage{soul}
\usepackage[utf8]{inputenc}

\usepackage[OT2, T1]{fontenc}
\usepackage[ukrainian, english]{babel}

\usepackage[normalem]{ulem}

\makeatletter
\newcommand{\thickhline}{%
    \noalign {\ifnum 0=`}\fi \hrule height 1.1pt
    \futurelet \reserved@a \@xhline
}
\makeatother

\title{
Revisiting Cross-Lingual Summarization: A Corpus-based Study and\\A New Benchmark with Improved Annotation
}

\author{
  Yulong Chen$^{1, 2}$ \quad
  Huajian Zhang$^{2}$ \quad
  Yijie Zhou$^{1}$ \quad
  Xuefeng Bai$^{2}$ \quad
  Yueguan Wang$^{2}$\\
  {\bf
  Ming Zhong$^{3}$\quad
  Jianhao Yan$^{2}$ \quad
  Yafu Li$^{2}$ \quad
  Judy Li$^{4}$ \quad 
  Michael Zhu$^{4}$\quad 
  Yue Zhang$^{2, 5}$ \Thanks{~Yue Zhang is the corresponding author.}
  }
  \\
  $^1$ Zhejiang University \quad $^2$ Westlake University \quad $^3$ UIUC \\
  $^4$ Sichuan Lan-bridge Information Technology Co., Ltd. \\
  $^5$ Westlake Institute for Advanced Study\\
  \emph{\href{mailto:yulongchen1010@gmail.com}{yulongchen1010@gmail.com}}\quad\quad\emph{\href{mailto:yue.zhang@wias.org.cn}{yue.zhang@wias.org.cn}}
  }
\begin{document}
\maketitle
\begin{abstract}
Most existing cross-lingual summarization (CLS) work constructs CLS corpora by simply and directly translating pre-annotated summaries from one language to another, which can contain errors from both summarization and translation processes.
To address this issue, we propose ConvSumX, a cross-lingual conversation summarization benchmark, through a new annotation schema that explicitly considers source input context.
ConvSumX consists of 2 sub-tasks under different real-world scenarios, with each covering 3 language directions.
We conduct thorough analysis on ConvSumX and 3 widely-used manually annotated CLS corpora and empirically find that ConvSumX is more faithful towards input text.
Additionally, based on the same intuition, we propose a 2-Step method, which takes both conversation and summary as input to simulate human annotation process.
Experimental results show that 2-Step method surpasses strong baselines on ConvSumX under both automatic and human evaluation.
Analysis shows that both source input text and summary are crucial for modeling cross-lingual summaries.
\end{abstract}

\section{Introduction}\label{sec:introduction}

With the advance in deep learning and pre-trained language models (PLMs)~\cite{devlin-etal-2019-bert, lewis-etal-2020-bart, DBLP:journals/jmlr/RaffelSRLNMZLL20}, much recent progress has been made in text summarization \cite{liu-lapata-2019-text, DBLP:conf/aaai/ZhongLX0022, chen2022unisumm}.
However, most work focuses on English (En) data~\cite{zhong-etal-2021-qmsum, gliwa-etal-2019-samsum, chen-etal-2021-dialogsum}, which does not consider cross-lingual sources for summarization~\cite{tacl_a_00520}. To address this limitation, cross-lingual summarization (CLS) aims to generate summaries in a target language given texts from a source language~\cite{zhu-etal-2019-ncls}, which has shown values to both academic and industrial communities~\cite{bai-etal-2021-cross, perez-beltrachini-lapata-2021-models}.

\begin{figure}[t!]
    \includegraphics[width=1.05\columnwidth]{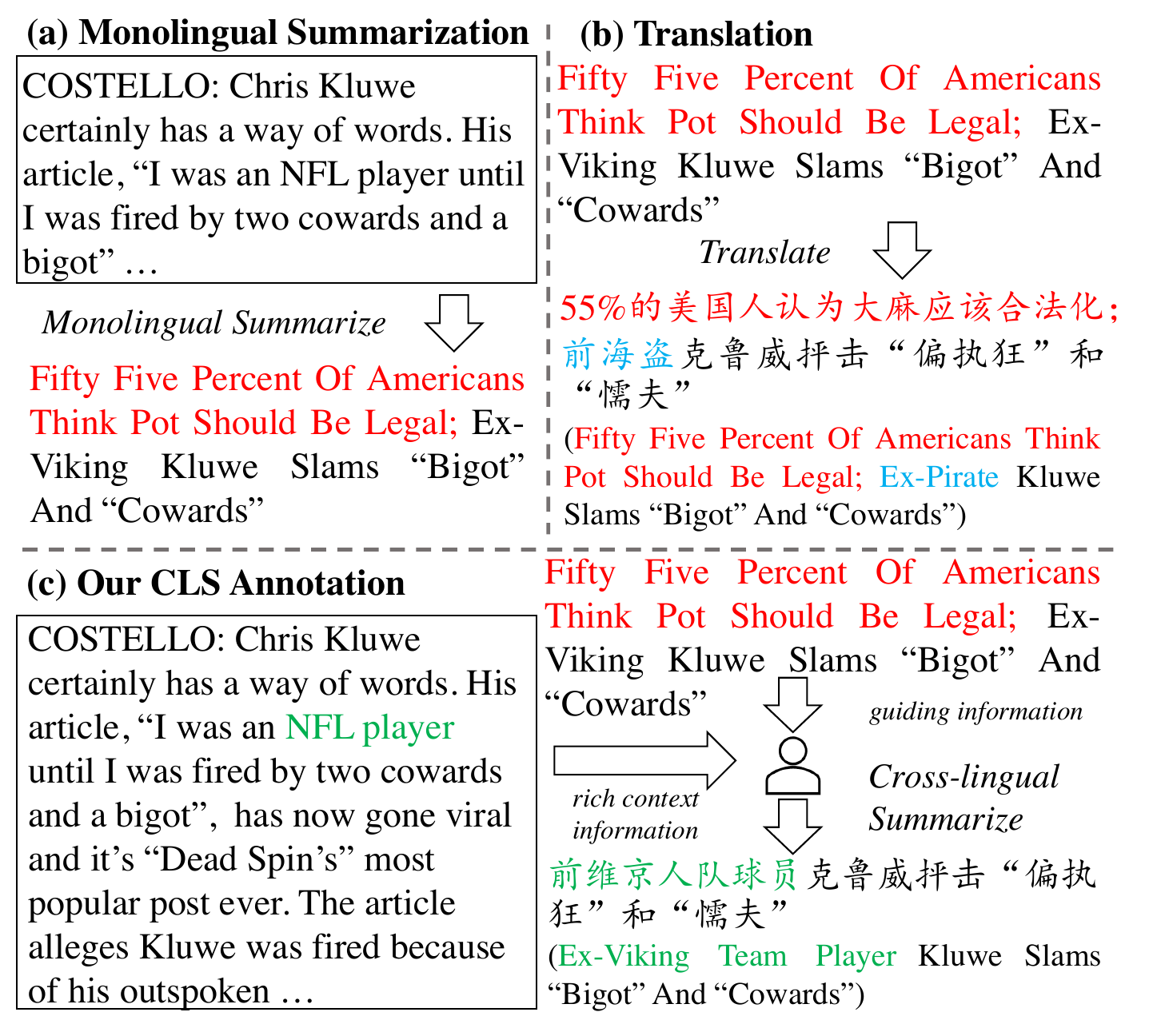}
    \centering
    \caption{
    An En-Zh summary from~\citet{wang2022clidsum} (best viewed in color).
    We compare ``\emph{pipeline}: (a)$\rightarrow$(b)'' annotation protocol and our annotation (c) protocol.
    Pipeline annotation results in errors from both summarization (\textcolor{red}{red}: unmentioned content/hallucination) and translation (\textcolor{cyan}{cyan}: incorrect translation) processes.
    To address this issue, we explicitly annotate target-language summaries with faithfulness rectification (\textcolor[RGB]{0,176,80}{green}) based on input context, with the guidance of mono-lingual summaries. 
    } 
    \label{fig:annotaton_methods}
\end{figure}

Most existing work~\cite{zhu-etal-2019-ncls, bai-etal-2021-cross,feng-etal-2022-msamsum} constructs CLS corpora by translating summaries from existing mono-lingual summarization datasets into other languages, which is de facto a ``\emph{pipeline}'' annotation protocol (first \emph{summarize}, then \emph{translate}) as shown in Figure~\ref{fig:annotaton_methods}.
However, such an annotation method can suffer from two major problems:
First, summaries from mono-lingual summarization corpora (summarization process) can contain errors~\cite{liu2022revisiting}, which are likely to be preserved in translated summaries.
For example, the English summary in Figure~\ref{fig:annotaton_methods}-(a) contains unmentioned content/hallucination (\textcolor{red}{red} text), which leads to the same discrepancy as in the translated summary (Figure~\ref{fig:annotaton_methods}-(b), \textcolor{red}{red} text).
Second, the translation process can further introduce errors, in particular for polysemous words.
For example, in Figure~\ref{fig:annotaton_methods}-(b), the term, ``\emph{Ex-Viking}'' (which refers to previous members of the Minnesota Vikings team), is mistakenly translated into ``\begin{CJK}{UTF8}{gbsn}\small{\textcolor{cyan}{前海盗}}\end{CJK}'' (which means ``\emph{ex-pirate/buccaneer}'').
To determine proper translation, it require more information beyond the scope of short summaries.

To qualitatively understand the above problems, we conduct human evaluation and error analysis on existing popular CLS corpora. Empirical results show that existing corpora suffer from the two aforementioned problems, containing significantly many hallucinations and factual errors.\footnote{The term \emph{error} later in this paper refers to errors that are hallucinations or can cause factual misunderstandings, except when otherwise specified.} 
In particular, we find that overall $20\sim67\%$ of summaries in CLS datasets contain errors, where $7\sim46\%$ and $13\sim47\%$ of summaries suffer from summarization and translation processes, respectively.
This suggests that the pipeline protocol, which is widely used in CLS research, can result in low-quality data and negatively impact the validity of modeling research. In addition, fine-grained error analysis shows that $55.6\sim89.1\%$ of translation errors can be resolved with the help of input context.

Motivated by the above findings and to address this issue, we propose the protocol that cross-lingual summaries should be sourced from original input text where mono-lingual summaries can serve as a quick review for salient information.
With this concept, we annotate cross-lingual summaries ($S^{tgt}$) by relying on source text ($D^{src}$) and source-language summaries ($S^{src}$) as shown in Figure~\ref{fig:annotaton_methods}-(c).
Such an annotation protocol brings three advantages: 
First, compared with translation only given $S^{src}$, rich context information from $D^{src}$ helps annotators to disambiguate word senses and comprehend $S^{src}$ accurately, e.g., ``\begin{CJK}{UTF8}{gbsn}\small{\textcolor[RGB]{0,176,80}{前维京人队}}\end{CJK}'' (which means ``\emph{ex-viking team player}'') in Figure~\ref{fig:annotaton_methods}-(c);
Second, $D^{src}$ is more reliable and can provide ground-truth information to correct potential errors in $S^{src}$, e.g., \textcolor{red}{red} text in Figure~\ref{fig:annotaton_methods}-(a); 
Third, compared with writing $S^{tgt}$ only given $D^{src}$, $S^{src}$ can serve as supplement guidance to help annotators be aware of what should be involved in the summaries, ensuring that salient information in $S^{src}$ and $S^{tgt}$ is aligned.

Using CLS protocol, we build ConvSumX, a new benchmark to facilitate future CLS research.
ConvSumX focuses on conversational text in a few-shot setting.
Compared with monologue (e.g., news), conversational text is less explored yet is also practically useful in real-world scenarios~\cite{chen-etal-2022-cross}.
ConvSumX contains two sub-tasks, namely DialogSumX and QMSumX, based on two English conversation summarization datasets \textsc{DialogSum}~\cite{chen-etal-2021-dialogsum} and QMSum~\cite{zhong-etal-2021-qmsum}, respectively.
Each covers three language-directions, taking En as the source, and Mandarin (Zh), French (Fr) and Ukrainian (Ukr) as target languages.
We empirically compare different annotations using the pipeline protocol and our CLS protocol with human evaluation.
Analysis shows that by considering input context, our protocol can significantly reduce annotation errors, suggesting ConvSumX is a high-quality benchmark in terms of cross-lingual faithfulness.

Based on the same intuition that $D^{src}$ and $S^{src}$ can serve as a critical complement to each other, we propose a 2-Step framework for CLS, which fine-tunes a multi-lingual PLM using concatenated $S^{src}$ and $D^{src}$ as input, and $S^{tgt}$ as output.
Experimental results show that our conceptual framework yields surprisingly better performance over strong baselines on ConvSumX.
Analysis and human evaluation show that our method can effectively generate more faithful cross-lingual summaries in a low-resource setting, and verify that source input text and summaries are supplementary to each other in modeling cross-lingual summaries.

To summarize, our contributions are the following:
\begin{itemize}[itemsep=0pt, topsep=0pt, parsep=0pt]
    \item[1.] We systematically review the pipeline annotation protocol and show that such a protocol can result in low-quality data (\S~\ref{sec:analysis_existing_cls_corpora});
    \item[2.] We propose the concept that CLS should be sourced from both source input text and source-language summaries and under our protocol, we present ConvSumX benchmark (\S~\ref{sec:convsumx}), where QMSumX is the first query-focused CLS dataset.
    \item[3.] Under the same concept, we propose a simple yet effective 2-Step framework for CLS (\S~\ref{sec:models}), which demonstrates the necessity of both source input text and mono-lingual summary for CLS modeling.
\end{itemize}
We release ConvSumX at \url{https://github.com/cylnlp/ConvSumX}.





\section{Analyzing Existing CLS Corpora}\label{sec:analysis_existing_cls_corpora}
We conduct a corpus-based study on existing popular human-annotated CLS corpora, namely NCLS, XSAMSum and XMediaSum, covering both monologue and dialogue texts.


\paragraph{NCLS}~\cite{zhu-etal-2019-ncls} is the first large cross-lingual news summarization corpus, which is constructed by automatically translating existing mono-lingual summarization datasets and using a round-trip strategy with human post-editing on test sets.

\paragraph{XSAMSum} and \textbf{XMediaSum} are both from \textsc{ClidSum}~\cite{wang2022clidsum}, where they manually translate summaries from two English dialogue summarization datasets, namely SAMSum~\cite{gliwa-etal-2019-samsum} and MediaSum~\cite{zhu-etal-2021-mediasum}, into Mandarin and German.


\subsection{Error Analysis on \emph{Pipeline} Annotation}
\label{sec:error_analysis_on_pipeline_annotaion}
Since all 3 corpora have the task of summarizing English (En) documents into Mandarin (Zh) summaries, we perform human evaluation on this language direction.
For each corpus, we randomly extract $100$ instances from its training and testing sets, respectively, resulting in a total of $600$ instances to evaluate.
Each instance consists of English document ($D^{en}$) and summary ($S^{en}$), and Mandarin summary ($S^{zh}$).

We invite two expert translators, who are native in Mandarin and professional in English, as our judges and ask them to first evaluate whether the $S^{zh}$ contains errors or not, by evaluating the $S^{zh}$ against $D^{en}$ (IAA\footnote{We measure Inter-Annotator Agreement (IAA) by calculating their Pair-wise Cohen kappa score on $60$ quiz instances.}: $0.67$, substantial agreement).
If $S^{zh}$ is found errors, the judges are asked to identify where such errors come from (IAA: $0.80$, substantial agreement).
Specifically, if this error is also found in $S^{en}$, we regard that it is caused by the mono-lingual summarization process; 
if this error is only found in $S^{zh}$ but not in $S^{en}$, we regard that it is caused by the translation process.
In this process, we only focus on factual errors, and minor syntax errors are ignored.

Table~\ref{tab:pilot_study} shows the evaluation result.
Overall, we see that all CLS corpora show high error frequencies ($20\sim67\%$), indicating existing CLS can be less accurate.
In particular, all mono-lingual summarization annotation contains errors ($7\sim46\%$), which are preserved in the CLS corpora.
Moreover, the cross-lingual annotation process can invite more errors ($13\sim47\%$).
This verifies our assumption that the pipeline annotation protocol, which ignores valuable input context, can lead to poor data quality.

In particular, NCLS contains the most errors, which can be because in addition to the different quality of their original mono-lingual summaries, $S^{zh}$ in NLCS are automatically translated by MT systems. 
Although human post-editing is conducted on the test set, factual errors are still frequent in the test set compared with the training set.
This can be because their post-editing focuses on poor fluency and translationese, while correcting factual errors or hallucinations requires information from the source text, which is not presented to human editors.
In addition, the averaged number of words in NCLS is much larger than in XMediaSum and XSAMSum,\footnote{Avg. token. length in English summaries: NCLS ($55.2$) XMediaSum ($14.4$), XSAMSum($20.3$).} making translation more difficult.

\begin{table}[t]
    \setlength\tabcolsep{2.5pt}
    \renewcommand{\arraystretch}{1.15}
    \centering
    \small
    \begin{tabular}{lr|ccc}
    \thickhline
        \multicolumn{2}{c|}{\textbf{Corpora}} & \textbf{Overall} & \textbf{Summ.}  & 
        \textbf{Trans.}\\
        \thickhline
        \multirow{2}{*}{NCLS}  &  Train & $67$ & $46$ & $47$ \\
        & Test & $60$ & $36$ &$40$ \\
        \hline
        \multirow{2}{*}{XMediaSum}  & Train & $27$ & $11$ & $19$ \\
        & Test & $27$ &$10$ & $18$ \\
        \hline
        \multirow{2}{*}{XSAMSum}  & Train & $35$ & $13$ & $23$ \\
        & Test & $20$ &\ \ $7$ & $13$ \\
    \thickhline

    \end{tabular}
    \caption{
    Error analysis on 3 CLS corpora. 
    We randomly sample 100 instances from the training and test sets, respectively, and count the number of factual errors.
    Summ. indicates that the mono-lingual summarization process contains errors, which are preserved in cross-lingual summaries;
    Trans. denotes that the errors in cross-lingual summaries are caused by the translation process. 
    One sample can have multiple errors.
    }

    \label{tab:pilot_study}
\end{table}


The major contradiction between frequent errors according to our analysis and the high data quality reported by \cite{zhu-etal-2019-ncls} and \cite{wang2022clidsum} can be explained by different reference sources, where our results show that these datasets have limitations in the choice of source for reference.
For example, when only given $S^{en}$ (``\emph{Fifty Five Percent... Ex-Viking...}'') as reference, an $S^{zh}$ (``\begin{CJK}{UTF8}{gbsn}\small{55\%的美国人...前海盗}\end{CJK}'') can be considered as a correct translation (Figure~\ref{fig:annotaton_methods}-b).
However, when evaluated against $D^{en}$, $S^{zh}$ is considered to have hallucination errors (``\begin{CJK}{UTF8}{gbsn}\small{55\%的美国人}\end{CJK}(\emph{fifty five percent...})'') and impropoer translation (``\begin{CJK}{UTF8}{gbsn}\small{前海盗}\end{CJK}(\emph{ex-pirate})'', which should have been translated into to ``\begin{CJK}{UTF8}{gbsn}\small{前维京人队员}\end{CJK}(\emph{ex-viking team member})'').

\begin{table}[t]
    \setlength\tabcolsep{3pt}
    \renewcommand{\arraystretch}{1.15}
    \centering
    \small
    \begin{tabular}{lr|cccccc}
    \thickhline
        \multicolumn{2}{c|}{\multirow{2}{*}{\textbf{Corpora}}} &  \multicolumn{6}{c}{\textbf{Translation Errors}}\\
        & & \textbf{W.S.}  & \textbf{Ter.} & \textbf{C.} &\textbf{S.R.} & \textbf{Oth.} & \textbf{All}\\
        \thickhline
        \multirow{2}{*}{NCLS}  &  Train & $25$ & $6$ & $2$ & $4$ &$12$ & $49$\\
        & Test & $23$ & $5$  & $5$ & $8$ &\ \ $5$ & $46$\\
        \hline
        \multirow{2}{*}{XMS}  & Train & \ \ $8$ &$3$ & $1$ & $3$ &\ \ $8$ & $23$ \\
        & Test & \ \ $5$ & $3$ &$0$ &$3$ &\ \ $8$ & $19$ \\
        \hline
        \multirow{2}{*}{XSS}  & Train &  \ \ $9$ & $5$ & $4$ &$4$ & \ \ $5$ & $27$ \\
        & Test &\ \ $4$ & $1$ & $1$ & $2$ & \ \ $5$ & $13$\\
    \thickhline
    \end{tabular}
    \caption{
    Fine-grained categorization of translation errors. Here we report the error count of each type.
    W.S, Ter., C., S.R., and Oth. stand for Word Sense, Terminology, Coreference, Sentence Relation, and Others.
    Note that one summary can have multiple errors.}
    \label{tab:pilot_study_error_analysis}
\end{table}

\subsection{In-depth Analysis on Translation Errors}
\label{sec:In-depth_analysis_on_translation_errors}
To further understand why directly translating English summaries can invite so many errors, we perform an error analysis on summaries containing translation errors and categorize them.
In particular, the two judges first identify whether the translation error can be resolved by considering the input context, or not, assuming that the errors can be caused by lacking input context (e.g., polyseme translation), and other translation errors (e.g., inconsistent translation).
We categorize the former error types based on their linguistic typologies (avg. IAA: $0.62$, substantial agreement):
\begin{itemize}[itemsep=0pt, topsep=0pt, parsep=0pt]
    \item[] \textbf{Word Sense (W.S.)}: the translation of a word/phrase is incorrect under source input context.
    \item[] \textbf{Terminology (Ter.)}: the translation of a word/phrase can be semantically correct but is improper in source input domains.
    \item[] \textbf{Coreference (C.)}: the translation of coreference expressions refer to incorrect objectives.
    \item[] \textbf{Sentence Relation (S.R.)}: The relation between two sentences/clauses is induced incorrectly or the translation of a sentence is incorrect because of misunderstanding the inter-relation/structure of a sentence.
    \item[] \textbf{Others (Oth.)}: simple errors such as typos or less accurate translation.
\end{itemize}


Table~\ref{tab:pilot_study_error_analysis} presents the error types and their error counts.
First, we see that errors (W.S, Tem., C. and S.R. together: $8\sim41$) caused by lacking input context are more than other translation errors (Oth.: $5\sim12$).
This further suggests the necessity of considering input text when annotating CLS corpora.
In addition, word sense sees overall most errors ($26.32\sim51.02\%$, avg. $41.81\%$), which is in line with the intuition that lacking context can mostly lead to word sense ambiguity.
Moreover, all categories see error instances, suggesting that such problematic summaries can confuse humans at multiple levels of language understanding.

Appendix~\ref{appendix:human} shows detailed information about our judges and Appendix~\ref{appendix:analysis_and_cases_of_translation_errors} shows cases of different translation error types and their analysis.

\section{ConvSumX}\label{sec:convsumx}
To address the aforementioned issues in pipeline annotation, we propose ConvSumX with a new annotation protocol, focusing on \emph{few-shot} CLS.
ConvSumX contains two cross-lingual summarization scenarios, namely daily dialogue summarization, and query-based summarization, covering 3 language directions: En2Zh, En2Fr and En2Ukr.



\subsection{Data Source}\label{sec:data_source}
We choose \textsc{DialogSum}~\citep{chen-etal-2021-dialogsum} and QMSum~\citep{zhong-etal-2021-qmsum} for ConvSumX by considering their potential to build real-world applications, and annotating their test and dev sets.

\paragraph{\textsc{DialogSum}} \textsc{DialogSum}~\cite{chen-etal-2021-dialogsum} is a real-life scenario dialogue summarization dataset, including various types of task-oriented dialogues.

\paragraph{{QMSum}} QMSum~\cite{zhong-etal-2021-qmsum} is a query-based meeting summarization dataset, covering the academic, product and committee domains.
We select data from academic and product for annotation.

\subsection{Annotation}
As discussed in \S~\ref{sec:analysis_existing_cls_corpora}, the final quality of CLS corpora can be influenced by both summarization process and translation process, most of which can be resolved with the information from input documents.
Therefore, instead of merely focusing on summaries in source languages, we ask annotators to write summaries in target languages ($S^{tgt}$) directly by considering both input documents ($D^{src}$) and pre-annotated summaries ($S^{src}$).
We refer to our protocol as CLS protocol.



We take English as the source language and choose Mandarin, French and Ukrainian as target languages because they are from different language families, and have different morphological variations and syntactic structures, with the potential to benefit other languages in their families.
We invite expert translators, who are native in target languages and professional in English, as our annotators (Appendix~\ref{appendix:human}).
We ask annotators to first comprehend $D^{src}$, and then write $S^{tgt}$ with the help of $S^{src}$.
In addition to the standard annotation criteria of \textsc{DialogSum} and QMSum, we ask our annotators specifically pay attention to the following aspects featuring the CLS:

\begin{itemize}[itemsep=0pt, topsep=0pt, parsep=0pt]
    \item Cross-lingual Consistency: Although being in different languages, the core semantic information of $S^{tgt}$ should be consistent with $D^{src}$, in particular for polysemous words or phrases.
    \item Language Style and Terminology: Annotators should write $S^{tgt}$ in the same language style of $S^{src}$, and use proper terminologies in some certain domains, such as academic meetings.
    \item Translationese: The annotated summaries should be natural in the target languages.
\end{itemize}

For QMSum, annotators are additionally asked to write a query in target languages ($Q^{tgt}$) with the help of the query in source language ($Q^{src}$), where $Q^{tgt}$ and $S^{tgt}$ form a QA pair.

\begin{table}[t]
    \renewcommand{\arraystretch}{1.15}
    \centering
    \small
    \begin{tabular}{lr|rr}
    \thickhline
        \multicolumn{2}{c|}{\textbf{Corpora}} & \textbf{Summ.} &\textbf{Query} \\
        \thickhline
        \multirow{2}{*}{\textsc{DialogSum}}  &  Dev & $34/500$ & $-$\\
        & Test & $21/500$ & $-$ \\
        \hline
        \multirow{2}{*}{QMSum}  & Dev & $33/199$ & $7/199$ \\
        & Test &\ \ $11/209$ & $0/209$\\
    \thickhline

    \end{tabular}
    \caption{
    Error analysis on QMSum and \textsc{DialogSum}.
    we show the number of error summaries/data size.
    }
    \label{tab:pre_annotation}
\end{table}


Before annotation, we ask each annotator to label training samples ($10\%$ of each dataset) until all annotated instances meet our requirements.
After annotation, each instance is reviewed by an editor, who is also an expert translator.
Editors are asked to first read the annotated summary to identify whether it is natural and readable in target languages, and then evaluate it against source input document to identify whether there are any factual errors.
If any errors are found, we ask the corresponding annotator to re-annotate the whole batch and repeat this checking and re-annotation process until all summaries are correct.
As mono-lingual summarization process can also contain large errors (\S~\ref{sec:error_analysis_on_pipeline_annotaion}), we additionally require annotators to modify English summaries/queries if any errors are found. 
Table~\ref{tab:pre_annotation} presents the percentage of summaries that contain errors in the original datasets.

Finally, we split the original dev sets into our new training and dev sets and keep the test set unchanged (DialogSumX: $400/100/500$  and QMSumX: $157/40/209$).

\begin{table}[]
    \setlength\tabcolsep{2.5pt}
    \renewcommand{\arraystretch}{1.15}
    \centering
    \small
    \begin{tabular}{lr|ccc}
    \thickhline
        \multicolumn{2}{c|}{\textbf{Corpora}} & \textbf{Overall} & \textbf{Summ.}  & 
        \textbf{Trans.}\\
        \thickhline
        \multirow{2}{*}{DialogSumX}  & T+D & \ \ $2$ &\ \ $0$ & \ \ $2$ \\
        & Test & \ \ $0$ &\ \ $0$ & \ \ $0$\\
        \hline
        \multirow{2}{*}{QMSumX}  & T+D &\ \ $2$ & \ \ $0$ &\ \ $2$\\
        & Test &\ \ $1$ &\ \ $0$ & \ \ $1$ \\
        \hline
        \multirow{2}{*}{DialogSum-P}  & T+D & $16$ &\ \ $9$ &\ \ $9$ \\
        & Test & $11$ &\ \ $5$ & \ \ $7$ \\
        \hline

        \multirow{2}{*}{QMSum-P}  & T+D & $31$ & $19$ & $18$ \\
        & Test & $19$ &\ \ $9$ & $13$ \\
    \thickhline

    \end{tabular}
    \caption{
    Comparison between CLS and pipeline annotation protocols.
    We count the number of different errors on $100$ instances, respectively.
    T+D: Training and Dev sets, which are the original dev set.
    }
    \label{tab:comparison_convsumx}
\end{table}

\begin{table*}[t]
    \setlength\tabcolsep{5pt}

    \renewcommand{\arraystretch}{1.15}
    \centering
    \small
    \begin{tabular}{l|ccccccc}
    \thickhline
        \textbf{Corpora} & \textbf{Domain} & \textbf{Lan. Direct} & \textbf{Annotation} &  \textbf{$D^{src}$}& \textbf{$S^{src}$} &\textbf{$S^{tgt}$} & \textbf{\% E.}  \\
    \thickhline
        En2ZhSum & News & En2Zh & $D^{src}\rightarrow S^{src}\leadsto S^{tgt}$ & 755.0 & 55.2 & 96.0 & 33.5 \\
        Zh2EnSum & News & Zh2En & $D^{src}\rightarrow S^{src}\leadsto S^{tgt}$ & 103.7 & 17.9 & 13.7 & - \\
        En2DeSum & News & De2En & $D^{src}\rightarrow S^{src}\leadsto S^{tgt}$ & 31.0 & \ \ 8.5 &\ \ 7.5 & - \\

         \hline
          XSAMSum & Written chit-chat & En2Zh/De & $D^{src}\rightarrow S^{src}\rightarrow S^{tgt}$ & 83.9 & 20.3& 33.0/19.9 & 27.5/- \\
          XMediaSum & Interview & En2Zh/De & $D^{src}\rightarrow S^{src}\rightarrow S^{tgt}$  & 1555.4 & 14.4 & 30.0/14.8 &  27.0/-\\
    \hline
    DialogSumX &  Real-life dialog& En2Zh/Fr/Ukr & $\{D^{src}, S^{src}\}\rightarrow S^{tgt}$ & 131.9 & 19.9 & 53.0/22.0/17.3 & 1.0/-/- \\
    QMSumX & Q-F meeting & En2Zh/Fr/Ukr & $\{D^{src}, S^{src}\}\rightarrow S^{tgt}$  & 1916.2 & 63.5 & 114.4/72.1/49.9&  1.5/-/-\\
    \thickhline
    \end{tabular}    
    \caption{Statistics of ConvSumX and other human-crafted CLS datasets.  Lan. Direct: language direction. \#: averaged length. 
    $D^{src}$,$S^{src}$ and $S^{tgt}$ are text length.
    We calculate character length for Mandarin and token length for others.
    Q-f: Query-focused.
    \% E.: averaged sampled error rate.
    Both Zh2EnSum~\cite{zhu-etal-2019-ncls} and En2DeSum~\cite{bai-etal-2021-cross} are constructed using the same method of En2ZhSum \cite{zhu-etal-2019-ncls}. ``$\rightarrow$'': human annotation. ``$\leadsto$'': automatic generation with human post-editing. 
    }
     \label{tab:comparison}
\end{table*}

\subsection{Comparison between ConvSumX with \emph{Pipeline} Annotation Data}\label{sec:comparison_between_convsumx_with_pipeline_annotation}
To qualitatively compare CLS and pipeline annotation protocols in a fair setting (e.g., to remove the influence of different data sources), we additionally annotate instances using the pipeline approach, i.e., directly translating English summaries into Mandarin. 
We randomly sample $100$ instances from dev/test sets of \textsc{DialogSum} and QMSum, referring to them as DialogSum-P and QMSum-P, respectively.
Overall, we have $400$ instances to annotate and $800$ instances to evaluate.

These data are annotated by the same annotators, using the same quality control process as ConvSumX.
To avoid priori knowledge from input context for pipeline annotation, this process is conducted \emph{before} ConvSumX annotation.
Then, we perform human evaluation on those translated data and corresponding data in ConvSumX using the same method as described in \S~\ref{sec:error_analysis_on_pipeline_annotaion} in an anonymous way.
For ConvSumX, we take corrected English summaries as \emph{pseudo} translation for evaluation.
Table~\ref{tab:comparison_convsumx} shows the human evaluation results.

Consistent with our findings (\S\ref{sec:error_analysis_on_pipeline_annotaion}), DialogSum-P and QMSum-P contain errors ($11\sim31$) from both the summarization and translation processes.
In contrast, ConvSumX contains fewer errors ($0\sim2$),\footnote{All errors that we find in \S~\ref{sec:comparison_between_convsumx_with_pipeline_annotation} are further corrected in ConvSumX. The final ConvSumX, which is used for training and evaluating models in \S~\ref{sec:experimental_setup}, contains no errors that we can find.} indicating the necessity of our CLS annotation protocol.

\subsection{Characteristics of ConvSumX}
Table~\ref{tab:comparison} presents a comparison between ConvSumX and other CLS corpora, highlighting the unique features of ConvSumX. Firstly, ConvSumX is designed for spoken conversation summarization and encompasses two real-world scenarios. 
Notably, QMSumX is the first corpus addressing query-based CLS. Secondly, ConvSumX includes multiple languages from diverse families (French: Romance; Mandarin: Chinese; Ukrainian: Slavic; English: Germanic), positioning it as a valuable resource for studying cross-lingual generalization and language transfer.
Furthermore, ConvSumX is the pioneering benchmark for CLS research involving the low-resource language, Ukrainian. 
Last, ConvSumX is the first CLS benchmark that forsakes the pipeline annotation protocol, which is essentially different from all existing human-crafted corpora. 
The low error frequencies demonstrate its cross-lingual faithfulness.


\begin{figure*}
    \centering
    \includegraphics[width=0.85\textwidth]{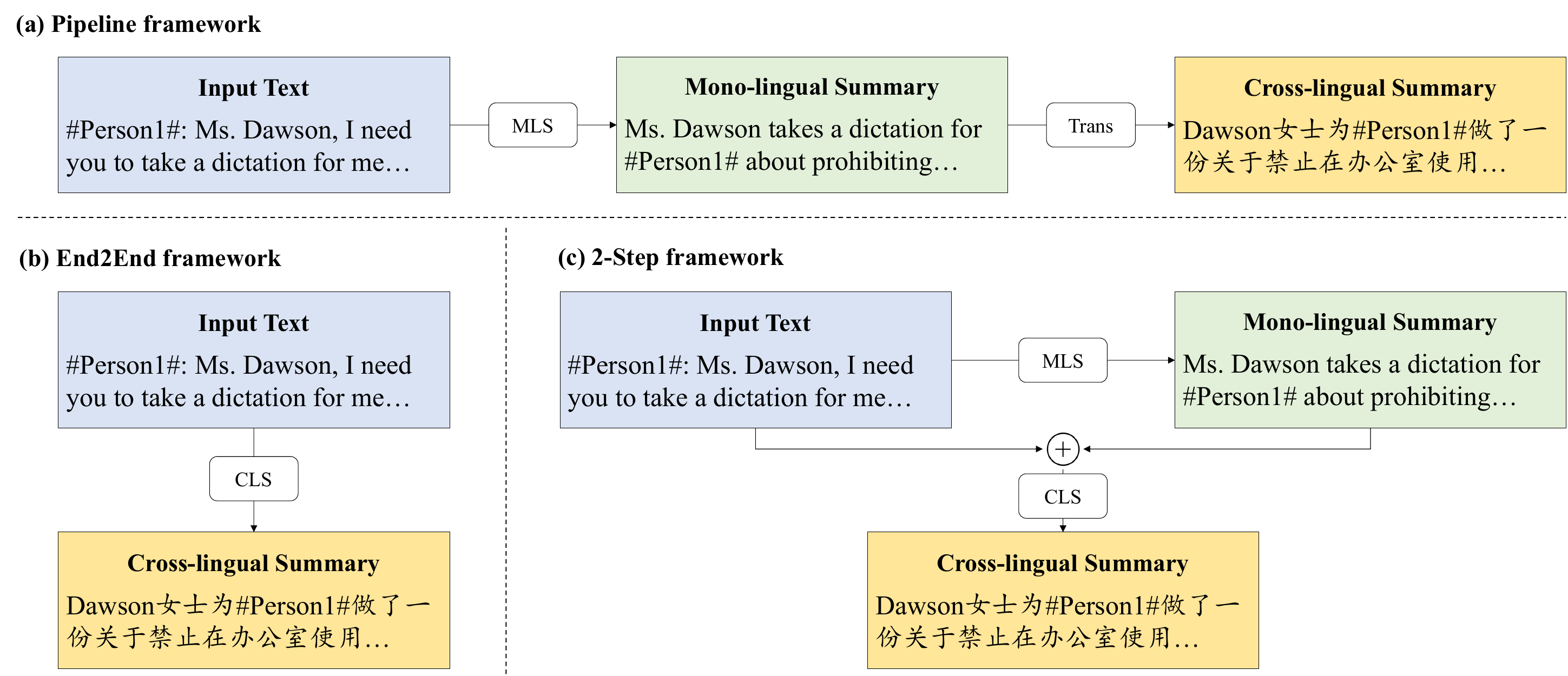}
    \caption{Illustration of pipeline method, end2end method, and our 2-Step method. MLS: mono-lingual summarizer; CLS: cross-lingual summarizer; Tans: translator.}
    \label{fig:cls_models}
\end{figure*}



\section{Method}\label{sec:models}
\subsection{Setting}
Generally, the task of \emph{{few-shot CLS}} is defined as:
given a source input text $D^{src}$, few-shot CLS is to generate a summary in a target language $S^{tgt}$ by learning a limited number of gold-annotated $\langle D^{src}, S^{tgt}\rangle$ data, with the help of external knowledge, which can be from mono-lingual summarization data, machine translation data and PLMs.

Specifically, for \emph{{query-focused CLS}}, the system is asked to generate  $S^{tgt}$ given $D^{src}$ with a query in the target language  $Q^{tgt}$.

\subsection{Models}
We evaluate two standard CLS baselines, namely pipeline method and End2End method, and propose a novel 2-Step framework, which differ from each other in the way the cross-lingual summary is generated.
Figure~\ref{fig:cls_models} summarizes the main difference between their workflows.

\paragraph{Pipeline Method} Previous work decomposes CLS into mono-lingual summarization and machine translation~\cite{zhu-etal-2019-ncls}, by deploying \emph{first-summarize, then-translate} (\emph{S-T}) or \emph{first-translate, then-summarize} (\emph{T-S}) strategies.

We compare with \emph{S-T} as it can benefit from large mono-lingual summarization and monologue translation data, while \emph{T-S} has been proven much worse~\cite{feng-etal-2022-msamsum} as both dialogue translation and non-English summarization data are very limited.
For QMSumX, we additionally translate 
$Q^{tgt}$ into $Q^{src}$ before mono-lingual summarization and translation, to which we refer as \emph{T-S-T}.


\paragraph{End2End Method} Previous work models the CLS task and has shown better performance on previous datasets compared with pipeline methods~\cite{zhu-etal-2019-ncls, xu-etal-2019-cross}.

We compare two End2End methods:
First, we directly fine-tune a multi-lingual model on $\langle D^{src},S^{tgt}\rangle$ (DialogSumX) and  $\langle \{Q^{tgt};D^{src}\},S^{tgt}\rangle$ (QMSumX), marked as E2E;
Second, inspired by \citet{bai-etal-2021-cross}, where an End2End model first generates mono-lingual summary and then cross-lingual summary in an auto-regressive way and shows good performance in few-shot setting, we fine-tune a multi-lingual model on $\langle D^{src},\{S^{src};S^{tgt}\}\rangle$ (DialogSumX) and  $\langle \{Q^{tgt};D^{src}\},\{S^{src};S^{tgt}\}\rangle$ (QMSumX), marked as E2M (M means mixed).

\paragraph{2-Step Method} Inspired by our data analysis (\S~\ref{sec:analysis_existing_cls_corpora}) that mono-lingual summary can help guiding salient information for cross-lingual summary, and generating proper translation requires information from source input text, we proposed a 2-Step method.
Conceptually, 2-Step is designed to simulate human annotation, where we ask an end2end model to generate $S^{tgt}$ given concatenated $S^{src}$ and $D^{src}$.
Compared with pipeline methods, 2-Step method can explicitly make use of information from source input.
Compared with End2End methods, 2-Step can focus on relevant information with the help of mono-lingual summaries.

Similarly, for QMSumX, we obtain the source language summaries by first translating $Q^{tgt}$ into $Q^{src}$ and then using mono-lingual summarizers.
During inference, we use model-generated source summaries as $S^{src}$, which are obtained using the same way of pipeline methods.

Note all individual models are in a seq2seq manner. The terms ``\emph{pipeline}'', ``End2End'' and ``2-Step'' are stated from the perspective between source input text and output cross-lingual summaries.

\begin{table*}[t]
    \setlength\tabcolsep{2.5pt}
    \renewcommand{\arraystretch}{1.15}
    \centering
    \small
    \begin{tabular}{l|cccc|cccc|cccc|cccc}
    \thickhline
\multirow{3}{*}{\textbf{Model}} &   \multicolumn{16}{c}{\textbf{DialogSumX}} \\
        &  \multicolumn{4}{c}{\textbf{\emph{En2Zh}}}
        & \multicolumn{4}{c}{\textbf{\emph{En2Fr}}}
        & \multicolumn{4}{c}{\textbf{\emph{En2Ukr}}}
        & \multicolumn{4}{c}{\textbf{\emph{Avg.}}}\\
        \cline{2-17}
          &     \textsc{R1} &   \textsc{R2} &  \textsc{RL}   & \textsc{BS} &   \textsc{R1} &   \textsc{R2} &  \textsc{RL}  & \textsc{BS} &   \textsc{R1} &   \textsc{R2} &  \textsc{RL}  & \textsc{BS} &   \textsc{R1} &   \textsc{R2} &  \textsc{RL}  & \textsc{BS}   \\
 
        \thickhline
        \emph{S-T} &
        46.32 & 24.08 & 39.51 & 78.36 &
        46.12 &23.66& \textbf{37.76} & 80.43 &
        \textbf{36.19} & 18.44 & \textbf{31.80} & \textbf{78.30} &
        42.88 & 22.06 & 36.36 & 79.03
        \\

        E2E & 
        41.33 & 20.14 & 34.74 &  76.66 &
        39.96 & 17.81 & 31.14 & 77.73 &
        31.42 & 14.61 & 26.95 & 76.33 &
        37.57 & 17.52 & 30.94 & 76.91
        \\
        E2M & 
        39.12 & 18.94 & 33.70 & 75.45& 
        39.51 & 16.96 & 30.92 & 77.33 &
        30.24 & 13.52 & 26.03 & 76.11 &
        36.29 & 16.47 & 30.22 & 76.30
        \\
        2-Step &
        \textbf{46.87} & \textbf{24.48} & \textbf{39.92} & \textbf{79.10} &
        \textbf{46.19} & \textbf{23.82} & 37.65 & \textbf{80.46} &
        36.05 & \textbf{18.46} &31.60 &78.24 &
        \textbf{43.04} & \textbf{22.25} & \textbf{36.39} & \textbf{79.27}
        
        \\
        \thickhline
        \multirow{3}{*}{\textbf{Model}} &   \multicolumn{16}{c}{\textbf{QMSumX}} \\
        &  \multicolumn{4}{c}{\textbf{\emph{En2Zh}}}
        & \multicolumn{4}{c}{\textbf{\emph{En2Fr}}}
        & \multicolumn{4}{c}{\textbf{\emph{En2Ukr}}}
        & \multicolumn{4}{c}{\textbf{\emph{Avg.}}}\\
        \cline{2-17}
          &     \textsc{R1} &   \textsc{R2} &  \textsc{RL}   & \textsc{BS} &   \textsc{R1} &   \textsc{R2} &  \textsc{RL}  & \textsc{BS} &   \textsc{R1} &   \textsc{R2} &  \textsc{RL}  & \textsc{BS} &   \textsc{R1} &   \textsc{R2} &  \textsc{RL}  & \textsc{BS}   \\
        \thickhline
        \emph{T-S-T} &
        31.89 & \ \ 7.82 & 22.03 & 68.45 &
        38.74 & 13.49 & 24.26 & 74.19 &
        20.15 & \ \ 5.55 & \textbf{14.44} & 71.57 & 
        30.26 &  \ \ 8.95 & 20.24  & 71.40
        \\
        E2E & 
        30.74 & \ \ 6.84 & 21.98 & 67.81 &
        35.81 & 11.38 & 22.24 & 72.96 & 
        16.76 & \ \ 4.52 & 12.22 & 69.54 &
        27.77 & \ \ 7.58 & 18.81 & 70.10
        \\
        E2M & 
        30.09 & \ \ 6.59 & 20.91 & 67.47 &
        32.51 & 10.01 & 20.66 & 70.90 &
        17.93 & \ \ 4.88 & 12.92 & 69.58 &
        26.84 & \ \ 7.26 & 18.16 & 69.32 
        \\

        2-Step &
        \textbf{33.20} &\ \ \textbf{8.43} & \textbf{23.12} & \textbf{69.36}&
        \textbf{38.91} & \textbf{13.52} & \textbf{24.37} & \textbf{74.27} &
        \textbf{20.51} &\ \ \textbf{5.73} & 14.38 & \textbf{71.75} &
        \textbf{30.87} & \ \ \textbf{9.23} & \textbf{20.63} & \textbf{72.79}
        \\
        \thickhline

    \end{tabular}
    \caption{
    Main results on ConvSumX. \emph{S-T} and \emph{T-S-T}: pipeline methods that decompose CLS as mono-lingual summarization and translation tasks; E2E: End2End method that directly generates target summaries; E2M: End2End method that generates source summaries and target summaries sequentially; 2-Step: our method that first generates source summaries, and generates target summaries with mono-lingual summaries as guiding information.
    }
    \label{tab:main_results}
\end{table*}

\section{Experiments}
\label{sec:experimental_setup}

\paragraph{Metrics}
\label{sec:evaluation_metrics}
For automatic evaluation,
we use \textsc{Rouge}~\cite{lin-2004-rouge}\footnote{\url{https://github.com/csebuetnlp/xl-sum}} and \textsc{BERTScore}~\cite{DBLP:conf/iclr/ZhangKWWA20}\footnote{\url{https://github.com/Tiiiger/bert_score}}.
\textsc{Rouge} measures the $n$-gram overlap between generated and reference summaries.
\textsc{BERTScore} calculates the pariwise cosine similarity between \textsc{BERT}~\cite{devlin-etal-2019-bert} token embeddings of generated and reference summaries.
We report the $F$-1 scores of \textsc{Rouge-1} (\textsc{R1}), \textsc{Rouge-2} (\textsc{R2}), \textsc{Rouge-L} (\textsc{RL}) and \textsc{BERTScore} (\textsc{BS}).

\paragraph{Implementation Details}\label{sec:implementation_details}
For mono-lingual generation, we use \textsc{UniSumm}\footnote{\url{https://github.com/microsoft/UniSumm}} for model initialization, further pre-training it on original training sets of \textsc{DialogSum} and QMSum, and then prefix-tuning it on our few-shot training data.
For cross-lingual generation (MT or CLS), we use \texttt{mBART-large-50-many-to-many-mmt}\footnote{\url{https://huggingface.co/facebook/mbart-large-50-many-to-many-mmt}} for model initialization and then fine-tune it on our cross-lingual data.
All experiments are conducted on NVIDIA A100 GPU.
We conduct a hyper-parameter search for learning rate and batch size, from [1.5e-4, 1e-4, 5e-5, 3e-5, 1e-5] and [8, 16, 32, 64], and choose the best checkpoint based on \textsc{R2} score on our few-shot dev sets.

\subsection{Main Results}
The main results on DialogSumX (\emph{DX}) and QMSumX (\emph{QX}) are shown in Table~\ref{tab:main_results}.
In general, we find that our 2-Step system achieves the best results in most languages and the best averaged results on both tasks.
In particular, 2-Step system outperforms pipeline method (\emph{S-T}) (avg. improvement: $0.19$ \textsc{R2} and $0.24$ \textsc{BS} scores on \emph{DX}; $0.61$ \textsc{R2} and $1.39$ \textsc{BS} scores on \emph{QX}).
 It also outperforms End2End models by a large margin (avg. improvement: $4.73\sim5.78$ \textsc{R2} and $2.36\sim2.79$ \textsc{BS} scores on \emph{DX}; $1.65$ \textsc{R2} and $2.69$ \textsc{BS} scores on \emph{QX}).
 Note that 2-Step system is additionally presented with source summary and input text information compared with E2E and \emph{S-T} systems.
Thus, the superiority of 2-Step demonstrates that the source document and source summary are crucial in modeling cross-lingual summaries, and are complementary to each other.

Moreover, \emph{S-T} outperforms End2End models.
The contradiction between our results and previous findings~\cite{bai-etal-2021-cross, chen-etal-2022-cross} can be explained by the fact that the summarizer and translator we use are much stronger and the error propagation problem is less severe.
Also, \emph{S-T} can benefit from our high-quality parallel cross-lingual summary pairs ($S^{src}$ and $S^{tgt}$) as few-shot translation data, while previous work ignores such valuable data and only uses a fixed MT system without fine-tuning~\cite{zhu-etal-2019-ncls}.

All CLS systems perform better at En2Zh and En2Fr than En2Ukr.
The high performance on En2Zh and En2Fr can be explained by that both Zh and Fr are highly-rich resource data on which mBART-50 is pre-trained~\cite{tang-etal-2021-multilingual}, and mBART-50 can easily bridge the alignment between texts in Zh/Fr and En.
In contrast, Ukr is a low-resource language, on which the mBART-50 performs poorly.
All systems have higher performance on \emph{DX} compared with \emph{QX}, which is because \emph{QX} is more challenging w.r.t the task of query-based summarization for long text and more extreme few-shot setting, and its domain is very different from mBART-50's pre-training data.

We notice that all models perform better on \emph{QX} En2Fr than En2Zh and En2Ukr.
A possible reason can be that \emph{QX} contains many professional in-domain words whose word sense can be multiple and very different from its general ones.
The sense of these words can be different lexical items, in particular for Zh or Ukr, which are typologically different from En~\cite{chen1989semantic,budzhak1998against}.
In contrast, Fr and En both use Latin script and are more similar in terms of morphology and lexicon rules~\cite{kirsner1984bilingual,pacton2008timing,fan2021beyond} compared with Zh and Ukr.
For example, ``\emph{discourse}'' can be mapped into ``\begin{CJK}{UTF8}{gbsn}\small{论文(academic paper)/讲述(talk)/...}\end{CJK}'' in Zh and ``\foreignlanguage{ukrainian}{diskusicq} \begin{CJK}{UTF8}{gbsn}\small{(discussion)/}\end{CJK}\foreignlanguage{ukrainian}{diskurs} \begin{CJK}{UTF8}{gbsn}\small{(linguistic discourse)}\end{CJK}'' in Ukr, while ``\emph{discours} \begin{CJK}{UTF8}{gbsn}\small{(discussion/linguistic...)}\end{CJK}'' in Fr.



We also conduct experiments on pipelined datasets, XSAMSum and XMediaSum (Appendix~\ref{appendix:clidsum}).  
Experimental results show that, with a fine-tuned translator, \emph{S-T} method outperforms best-reported systems on most tasks.
Moreover, 2-Step does not show better performance than \emph{S-T}, which can be because 2-Step systems are trained to only translate source summaries instead of comprehending source input text.
The high performance of \emph{S-T} emphasizes that cross-lingual summaries in those pipelined datasets do not rely on source input text, which can rather be a translation task.
This confirms our motivation that the pipeline annotation protocol has important limitations.

\subsection{Human Evaluation}\label{sec:human_evaluation}
To comprehensively understand CLS systems, we conduct human evaluations of the model outputs, as multi-dimensional assessment offers a more robust and holistic perspective~\cite{DBLP:conf/emnlp/Zhong0YMJLZJH22}.

Following previous work~\cite{kryscinski-etal-2019-neural, fabbri-etal-2021-summeval}, we evaluate generated summaries from the following dimensions:
\emph{Fluency} evaluates the quality of generated sentences, including grammar and whether it is natural; \emph{Coherence} evaluates the collective quality of generated summaries; \emph{Relevance} evaluates the importance of information in generated summaries; \emph{Consistency} evaluates factual alignment between generated summaries and source input texts.
We randomly extract $50$ summaries from \emph{S-T} and 2-Step outputs on ConvSumX for each language, and ask native speakers to give scores from $1$ to $5$.
Higher scores indicate higher qualities.

The result is shown in Table~\ref{tab:human_eval}.
Generally, all metrics see low scores, suggesting the challenge of few-shot CLS.
Both models see higher scores on \emph{DX} compared with \emph{QX}, which is consistent with our automatic evaluation.
Compared with \emph{S-T}, 2-Step achieves similar Relevance scores on all tasks.
This is because the input source summary for both models is identical, thus the information in it is the same.
However, 2-Step achieves higher Fluency, Coherence, and Consistency scores, which justifies our assumption that source input text information is critical, in particular for consistency.

We present case study of model outputs in Appendix~\ref{appendix:case_study}.

\begin{table}[t]
    \setlength\tabcolsep{2.5pt}
    \renewcommand{\arraystretch}{1.15}
    \centering
    \small
    \begin{tabular}{lr|cccc|cccc}
    \thickhline
    \multicolumn{2}{c|}{\multirow{2}{*}{\textbf{Model}}} & \multicolumn{4}{c|}{\textbf{\emph{DX}}} & \multicolumn{4}{c}{\textbf{\emph{QX}}} \\
    &&
    \makebox[0.01\textwidth][c]{\emph{F.}} & \makebox[0.01\textwidth][c]{\emph{Coh.}} &\makebox[0.01\textwidth][c]{\emph{Con.}} &\makebox[0.01\textwidth][c]{\emph{R.}} & \makebox[0.01\textwidth][c]{\emph{F.}} & \makebox[0.01\textwidth][c]{\emph{Coh.}} &\makebox[0.01\textwidth][c]{\emph{Con.}}
    &\makebox[0.01\textwidth][c]{\emph{R.}} \\
    \thickhline
    \multirow{3}{*}{\emph{S-T}} 
    &En2Zh & 2.60 & 2.87 &	2.27 &	3.30  & 2.10&	2.15 & 1.95	& 2.25\\
    &En2Fr & 3.23	& 4.43	& 3.37 &	2.50 & 2.85	&3.65&	1.60	&1.35\\
    &En2Ukr & 3.90 & 3.57 & 3.20 & 3.20 & 3.30 & 3.25  & 2.90 & 3.00 \\
    \hline
    \multirow{3}{*}{2-S} 
    &En2Zh & 2.90 & 3.00 & 2.50 &3.30 & 2.40 & 2.45 & 2.20 & 2.45\\
    & En2Fr &  3.30 &	4.47& 3.47 & 2.50 & 3.00	& 3.65 &	1.90 &	1.50\\
    & En2Ukr &3.83 & 3.70 & 3.57 & 3.30 & 3.35 & 3.25 & 3.00 & 3.05 \\   
    \thickhline
    \end{tabular}
    \caption{
    \emph{F.}, \emph{Coh.}, \emph{Con.} and \emph{R.} are \emph{Fluency}, \emph{Coherence}, \emph{Consistency} and \emph{Relevance}. 2-S: 2-Step.
    Please note that the scores are not comparable between languages.
    }
    \label{tab:human_eval}
\end{table}

\section{Related Work
}\label{sec:related_work}
\paragraph{CLS Corpora}\label{sec:cls_corpora}
Existing CLS corpora construction can be categorized into two main protocols.
1) Pipeline annotation: translating summaries from MLS corpora into other languages and;
2) Automatic alignment: aligning summaries and input texts of different language versions.

\citet{zhu-etal-2019-ncls} construct the first large-scale CLS dataset by automatically translating mono-lingual summaries using MT systems with a round-trip strategy and manual post-editing on test sets.
\citet{bai-etal-2021-cross} construct an En2De dataset using the same method.
\citet{feng-etal-2022-msamsum} automatically translate summaries from SAMSum~\cite{gliwa-etal-2019-samsum} into Russian, De and Zh. \citet{wang2022clidsum} manually translate summaries from SAMSum~\cite{gliwa-etal-2019-samsum} and MediaSum~\cite{zhu-etal-2021-mediasum} into De and Zh.
Different from them, we propose a new annotation protocol, which helps annotators to comprehend documents quickly and accurately.
To our knowledge, we are the first to address such human annotation issues for CLS research and present a new benchmark, ConvSumX. 

A different line of work constructs CLS datasets by linking different language versions of online articles, such as Wikipedia~\cite{perez-beltrachini-lapata-2021-models} and WikiHow~\cite{ladhak-etal-2020-wikilingua}.
Despite the cheap cost and large scale, there can be misalignment and hallucination problems.
For example, Wikipedia articles and their leading paragraphs (pseudo summaries) of the same person in different languages can contain different contents.
Also, such a method is limited to resources that contain multi-lingual data, which may not be available for all domains of interest, for example, the conversational text.


\paragraph{CLS Models}
Early work on CLS focuses on a pipeline paradigm by first summarizing, then translating, or vice versa.
However, due to the poor performance of early MT and summarization systems, such methods can often suffer from error propagation.
With the advance of deep learning and PLM technologies, recent work deploys end-to-end methods.
\citet{zhu-etal-2019-ncls}, \citet{xu-etal-2020-mixed}, \citet{bai-etal-2021-cross} and \citet{wang2022clidsum} propose multi-task learning or pre-training on large in-domain CLS, mono-lingual summarization and translation data.
Different from them, we propose a 2-Step method under the same concept of sourcing from source input text with the guidance of source summary, which is free of pre-training on large and thus can be easily adapted to other tasks and languages.

\section{Conclusion}\label{sec:conclusion}
We conducted data analysis on 3 typical corpora and showed that the pipeline annotation protocol suffers from errors from both the summarization and translation processes.
To address these issues, we proposed that cross-lingual summaries should be sourced from source input text.
Based on this principle, we annotated a more faithful CLS benchmark, ConvSumX by relying on both source-language texts and summaries.
Based on the same intuition, we proposed a 2-Step method that takes both source text and source summaries as input.
Experimental results showed that 2-Step method outperforms strong baselines on ConvSumX, demonstrating that both source-language texts and summaries are crucial in modeling cross-lingual summaries and are complementary to each other.
To our knowledge, we are the first to show that summary translation has limitations for CLS, giving a more faithful solution.

\section*{Limitations}

The limitation of this paper can be stated from three perspectives.
First, although using our CLS annotation protocol can label more faithful data, the annotation cost is higher because annotators need to comprehend the full source text instead of only the source summary.
Second, ConvSumX only covers 3 typical languages, while languages from different language families and have different morphology and lexical-/syntactic rules require further investigation.
Third, although the proposed 2-Step method is effective, we simply concatenate the source input text and mono-lingual summary at the token level as the model input but do not make further exploration.
We believe that more smart and sophisticated designs to integrate features from source input text and mono-lingual summary can further improve the CLS performance, which, however, we leave for future work.

\section*{Ethics Statement}
\paragraph{Data Usage and License} ConvSumX is based on two public English conversation summarization datasets, namely \textsc{DialogSum} and QMSum.
Both datasets are freely available online under the MIT license,  which has no constraint to academic use, modification, and further distribution.
We will follow the MIT license to make our data (annotated target summaries/queries and corrected English summaries/queries) freely available online.

\paragraph{Human Annotation } The construction of ConvSumX involves human annotation.
We hire $4$ expert translators as our annotators and editors for each target language.
The total cost is around $6,500$ USD, which applies to our annotation (including quiz annotation) and review.
The hourly salary is equal.
The total annotation time (including training annotation and editing) for Zh, Fr and Ukr is around $96$, $96$, and $120$ hours (according to our annotation cost/hourly salary).
Detailed information about our annotators/judges/editors can be found in Appendix~\ref{appendix:human}.

\paragraph{Content Safety} 
During our annotation, annotators are explicitly asked to not involve any personal/violent information and to write summaries strictly limited to the scope of source input text.
Also, if any violent or uncomfortable information is found in source input text, annotators are asked to report such issues.
All data are further reviewed by editors.
With careful checking and evaluation, ConvSumX (including source input text) contains no personal/violent content, and is safe to use.

\section*{Acknowledgement}
We thank reviewers from ACL2023 for their suggestions.
We extend our sincere and special thanks to our meta-reviewers for their indispensable and exceptional contributions. 
We also appreciate Ruochen Xu for insightful discussion and expert translators from Lan-bridge who have played a crucial role in the development of ConvSumX.
This work is funded by the Ministry of Science and Technology of China (grant No. 2022YFE0204900) and National Natural Science Foundation of China (grant NSFC No. 62161160339).

\bibliography{anthology,custom}
\bibliographystyle{acl_natbib}

\appendix
\clearpage


\section{Human Judges and Annotators}
\label{appendix:human}
For human evaluation in \S~\ref{sec:analysis_existing_cls_corpora}, we invite 2 expert translators as judges to conduct human evaluation and analysis of existing CLS corpora.
For cross-lingual summary annotation and mono-lingual correction (\S~\ref{sec:convsumx}), we invite 3 translators as our annotators and 1 as an editor to evaluate human annotation and model outputs (\S~\ref{sec:human_evaluation}) for each language direction.
Additionally, we invite one senior translator as the project manager to monitor the whole annotation progress.

All expert translators are from Lan-bridge, a qualified institution for translation service\footnote{Requirements for translation services: \url{https://www.iso.org/standard/59149.html}.}, recognized by the ISO\footnote{International Organization for Standardization: \url{https://www.iso.org/home.html}.}.
All annotators, editors and judges are native in the target language (i.e., Chinese, French or Ukrainian), and professional in English. 
They are competent in translation, linguistic research and related information processing. 
They also have a good understanding of the textual background of certain culture, technology and domain. 
Our annotators and editors either got graduate certificates in translation major or got graduate certificates in other fields but have more than 2 years of full-time professional experience in translating. 
Besides the above requirements, the manager has more than 5-year experience in multi-lingual translation projects that cover the language directions as described in this paper.



\begin{table}[t]
\setlength\tabcolsep{2.1pt}
\renewcommand{\arraystretch}{1.2}
    \centering
    \small
    \begin{tabular}{l|cccc|cccc}
    \thickhline
\multirow{3}{*}{\textbf{Model}} &   \multicolumn{8}{c} {\textbf{\emph{XSAMSum}}}  \\
&  \multicolumn{4}{c}{\textbf{\emph{En2Zh}}}
        & \multicolumn{4}{c}{\textbf{\emph{En2De}}}
        \\
        \cline{2-9}
          &     \textsc{R1} &   \textsc{R2} &  \textsc{RL}   & \textsc{BS} &   \textsc{R1} &   \textsc{R2} &  \textsc{RL}  & \textsc{BS}  \\
        \thickhline
        \emph{Summ-Trans$^*$} & 42.1 & 17.2 & 35.1 & \textbf{77.6} & \textbf{48.0} & \textbf{23.1} &\textbf{40.3} & \textbf{76.3}\\
        \emph{Trans$^*$-Summ} & 40.0 & 14.9 & 32.6 & 76.6 & 43.5 & 17.8  & 35.1 & 74.1 \\
        mBART$_\textsc{e2e}$ & 39.6 & 15.5&32.9 &76.6 & 43.4 & 17.6 & 35.9&74.0 \\
        mDBART$_\textsc{e2e}$ & -& -& -&- & - & -&- &-\\
        \hline
        \emph{S-T}$^*$\dag &36.0	& 12.3	&29.2& 74.1 
        & 43.3 & 16.5 & 34.5 & 73.4\\
        \emph{S-T}\dag & \textbf{43.8} & \textbf{18.7} & \textbf{35.9} &\textbf{77.6} & 46.2 & 20.0 & 37.6 & 75.1 \\
        2-Step\dag& 43.5 & \textbf{18.7} & 35.8 &\textbf{77.6} & 46.2 &20.2 & 37.6 & 75.1 \\
        \thickhline

\multirow{3}{*}{\textbf{Model}} & \multicolumn{8}{c}{\textbf{\emph{XMediaSum40K}}}\\
&  \multicolumn{4}{c}{\textbf{\emph{En2Zh}}}
        & \multicolumn{4}{c}{\textbf{\emph{En2De}}}
        \\
        \cline{2-9}
          &     \textsc{R1} &   \textsc{R2} &  \textsc{RL}   & \textsc{BS} &   \textsc{R1} &   \textsc{R2} &  \textsc{RL}  & \textsc{BS}  \\
        \thickhline
        \emph{Summ-Trans$^*$}  & 24.8 & 8.6 &22.2 & 66.8 & 23.9 & 9.9 & 21.2 & 62.0\\
        \emph{Trans$^*$-Summ}  & 24.1 & 8.2 & 21.4 & 65.9 & 20.9 &8.2 & 18.5 & 60.4\\
        mBART$_\textsc{e2e}$ & 23.8 & 7.8&21.0&66.0 & 20.6 &  7.7&18.2 &60.4  \\
        mDBART$_\textsc{e2e}$ & 28.7 & 11.1&25.7 &68.4  & 26.7& \textbf{12.1}&  \textbf{24.0}& \textbf{63.8}\\
        \hline
        \emph{S-T}$^*$\dag & 24.2 & 6.7 & 20.1 & 65.8&24.1	& 8.8	&21.0& 61.2 \\
        \emph{S-T}\dag& 29.6 & 11.1 & 25.9 & \textbf{68.5} & 27.3 &11.7 & \textbf{24.0} &63.6  \\
        2-Step\dag &  \textbf{29.7} & \textbf{11.2} & \textbf{26.0} & \textbf{68.5} &\textbf{27.4} & 11.8 & \textbf{24.0} &63.6  \\
        \thickhline

    \end{tabular}
    \caption{
    Experimental results on \textsc{ClidSum} dataset. 
    We show the best-reported pipeline and End2End method from \cite{wang2022clidsum}.
    \dag: our results.
    $^*$: translator that is not fine-tuned.
    For a fair comparison, we use BART-large~\cite{lewis-etal-2020-bart} for mono-lingual summarization and mBART as for cross-lingual generation.
    mDBART$_\textsc{e2e}$: mDialBART$_{E2E}$.
    }
    \label{tab:clidsum}
\end{table}

\section{Analysis and Cases of Translation Errors}
\label{appendix:analysis_and_cases_of_translation_errors}
As shown in Table~\ref{tab:error1} and Table~\ref{tab:error2}, we present cases of each error type as discussed in \S~\ref{sec:In-depth_analysis_on_translation_errors}.

In Table~\ref{tab:error1}, ``\emph{Sheen}'' refers to an actor, while annotators translate it into ``\begin{CJK}{UTF8}{gbsn}\small{高光}\end{CJK}/\emph{Highlight}'', and the term ``\emph{queer group}'' into ``\begin{CJK}{UTF8}{gbsn}\small{同性恋群体}\end{CJK}/\emph{gay group}''.
Although ``\emph{queer}'' has a meaning of ``\emph{gay}'', the proper translation should be ``\begin{CJK}{UTF8}{gbsn}\small{酷儿群体}\end{CJK}''.
Also, in the Coreference case, ``\emph{the date}'' refers to the day when ``\emph{go do groceries together}'', which is incorrectly translated into ``\begin{CJK}{UTF8}{gbsn}\small{聚会的日期}\end{CJK}/\emph{the date of party}''.
In Table~\ref{tab:error2} Sentence Relation, annotators confuse the meaning and relation between two sentences, and the translation is completely incorrect at the sentence semantic level.

All those translation cases together with summarization case (Figure~\ref{fig:annotaton_methods}) suggest the pipeline annotation can contain a large number of errors.

\input{error1.tex}

\input{error2.tex}


\section{Experiment on Pipelined Datasets}\label{appendix:clidsum}

We conduct experiments on two pipelined datasets, namely XSAMSum and XMediaSum from the \textsc{ClidSum} benchmark and compare our pipeline and 2-Step methods with best-reported systems from \cite{wang2022clidsum}:

\paragraph{\emph{Summ-Trans} Pipeline} They use BART($D_{all}$) for mono-lingual summarization~\cite{feng-etal-2021-language}, and Google Translate~\footnote{https://cloud.google.com/translate} for summary translation.

\paragraph{\emph{Trans-Summ} Pipeline} They use Google Translate to first generate cross-lingual dialogues, and then use mBART-50 to generate target language summaries.

\paragraph{mBART$_{E2E}$} They directly fine-tune an mBART-50 on cross-lingual $\langle S^{src},S^{tgt}\rangle$ pairs, which is also an E2E baseline in our \S~\ref{sec:models}.

\paragraph{mDialBART$_{E2E}$} They further pre-train an mBART-50 model using multiple tasks, including action filling, utterance permutation, mono-lingual summarization and machine translation, on MediaSum~\cite{zhu-etal-2021-mediasum}.

For more fair comparison, we fine-tune BART-large on corresponding mono-lingual data for mono-lingual summarization and fine-tune mBART-50 for translation and our 2-Step model. 
The result is shown in Table~\ref{tab:clidsum}.

We see that without fine-tuning, our pipeline method (\emph{S-T}$^*$) shows low performance.
However, equipped with fine-tuned mBART as translator, \emph{S-T} outperforms all previous methods on all tasks (e.g., \emph{S-T} outperforms the best-reported mDialBART on En2Zh XMediaSum by $1$ \textsc{R1}) except for the \emph{S-T} pipeline on En2De XSAMSum, which can be because that Google Translate is better than mBART-50 at En2De translation.
However, our 2-Step method, which is explicitly presented with source dialogue information and outperforms \emph{S-T} on \emph{ConvSum}, only shows comparable or even worse results compared with \emph{S-T} on XSAMSum and XMediaSum.
The contradiction of such model performance on \textsc{ClidSum} can be explained by that such pipelined datasets focus more on how to directly translate the mono-lingual summary, while adding more source dialogue information is less useful and sometimes harmful.

\begin{figure*}[th]
    \setlength{\belowcaptionskip}{-0.5cm}
    \setlength{\abovecaptionskip}{-0.0cm}
    \includegraphics[width=1.0\textwidth]{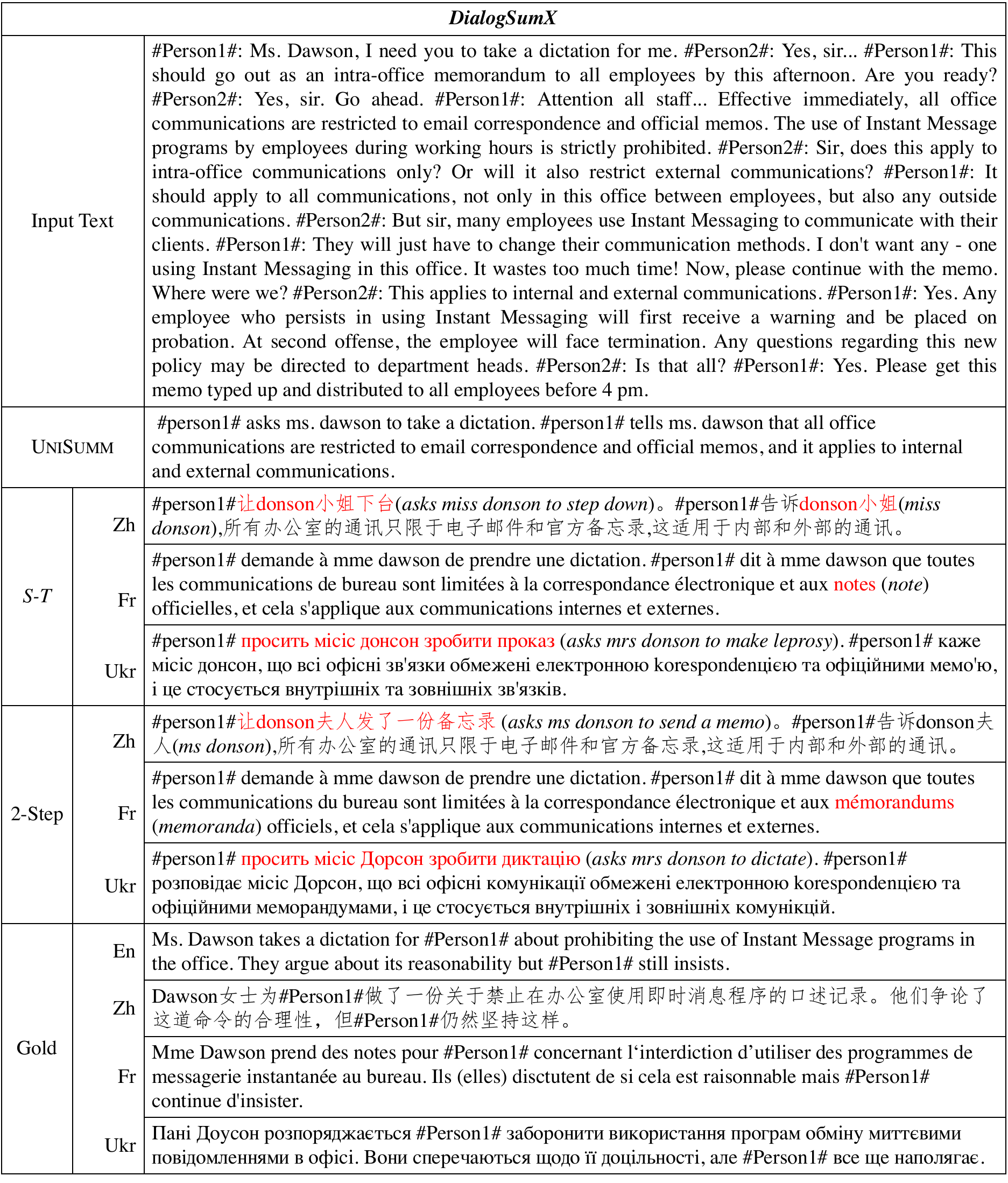}
    \centering
    \caption{
    Case (a): cross-lingual summaries generated by \emph{S-T} and 2-Step method, and human annotated summaries.
    We show their differences (not all errors) in \textcolor{red}{red} and
their English translation in bracketed {\it italics}.} 
    \label{fig:case_dialogsumx}
\end{figure*}

\begin{figure*}[th]
    \setlength{\belowcaptionskip}{-0.5cm}
    \setlength{\abovecaptionskip}{-0.0cm}
    \includegraphics[width=1.0\textwidth]{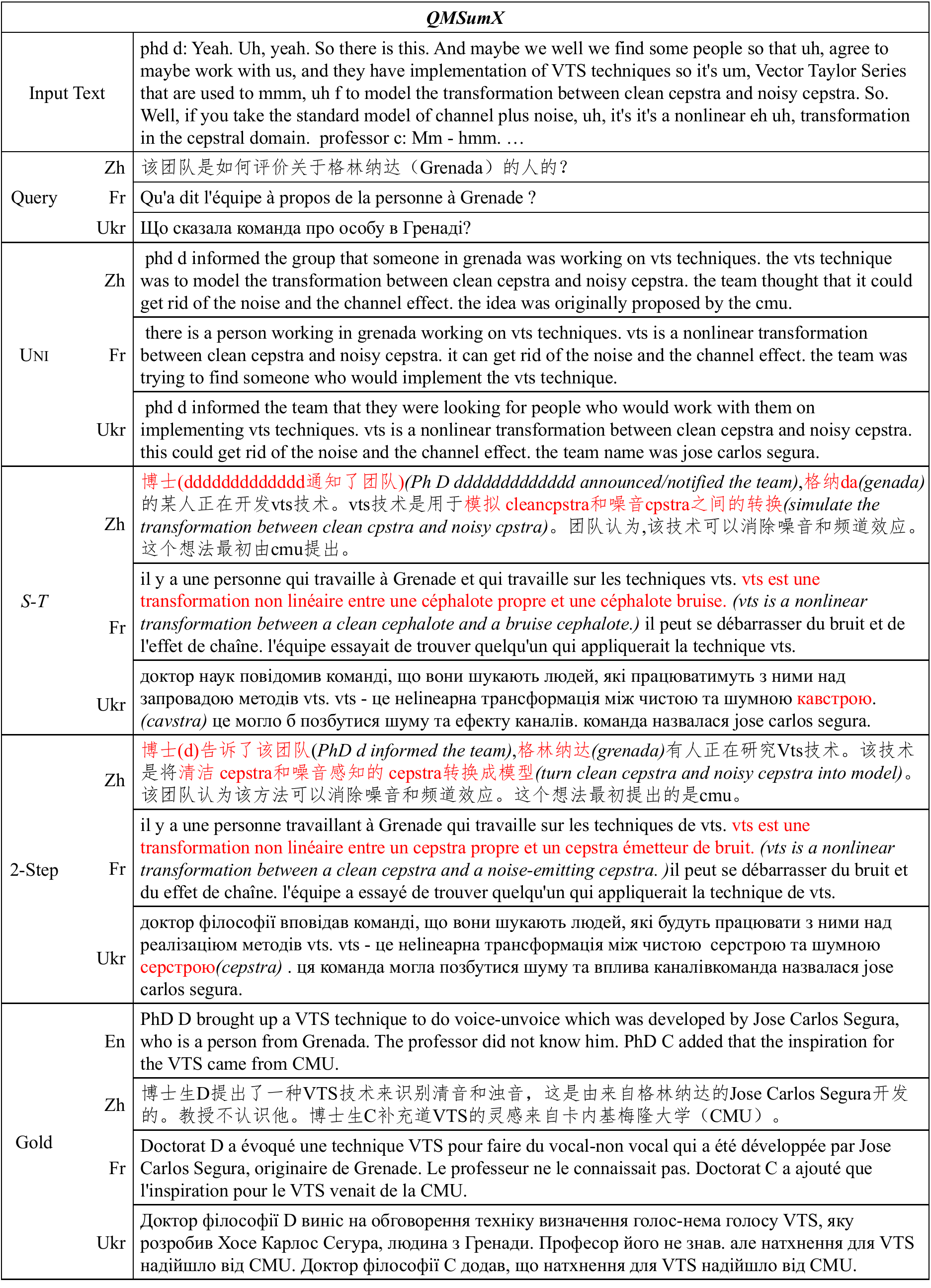}
    \centering
    \caption{
    Case (b): cross-lingual summaries generated by \emph{S-T} and 2-Step method, and human annotated summaries.
    We show their differences (not all errors) in \textcolor{red}{red} and
their English translation in bracketed {\it italics}.  }
    \label{fig:case_qmsumx}
\end{figure*}


\section{Case Study}\label{appendix:case_study}
To qualitatively demonstrate the advantage of 2-Step method, we present cases from \emph{S-T} and 2-Step on ConvSumX. 
Figure \ref{fig:case_dialogsumx} shows a sample from DialogSumX and Figure \ref{fig:case_qmsumx} shows another sample from QMSumX. 

As shown in  Figure \ref{fig:case_dialogsumx}, both summaries generated by  2-Step and \emph{S-T} contain errors (``\emph{donson}'', which should be ``\emph{dawson}'').
However, compared with \emph{S-T} (``\begin{CJK}{UTF8}{gbsn}\small{下台}\end{CJK}'', which means ``\emph{step down}'' or ``\emph{lose power}''), 2-Step method can improve the faithfulness of summaries (``\begin{CJK}{UTF8}{gbsn}\small{发了一份备忘录}\end{CJK}, which means ``\emph{send a memo}'').
Similarly, as shown in Figure~\ref{fig:case_qmsumx}, \emph{S-T} method suffers from generating unnatural language(e.g., a string of \emph{d}-s in {En2Zh} case) and it has trouble generating some not-commonly used words (e.g., the counterpart word of {\it cepstra} in 3 target languages), while 2-Step method can significantly ameliorate such problems. 

Moreover, we also find that 2-Step method not only focus on ``translating'' summaries from source language to target language, but also rewriting them by referring to the original input text. 
In the En2Zh example in Figure \ref{fig:case_qmsumx}, 2-Step method properly generates ``\begin{CJK}{UTF8}{gbsn}\small{告诉}\end{CJK}'' (which has a sense of ``\emph{tell}'') for the word ``\emph{inform}'', which is more natural considering its textual background. 
In contrast, \emph{S-T} method simply translates it into ``\begin{CJK}{UTF8}{gbsn}\small{通知}\end{CJK}'', which is more of the sense ``\emph{announce/notify}'', and is not natural in Mandarin.

These cases indicate that source-language texts are essential in cross-lingual summarization tasks, which further demonstrates our conclusion. 

\end{document}

%% file: error1.tex
\begin{table*}[h]
\renewcommand{\arraystretch}{1.5}
    \centering
    \small
    \begin{tabular}{l|p{13.5cm}}
    \thickhline
    \multicolumn{2}{c}{\textbf{Word Sense}}\\
    \thickhline
        Input Text  &
        BLITZER: \"Two and a Half Men\" minus one. Charlie Sheen has been fired. Warner Brothers Television terminated Sheen's contract with the sitcom hit today. CNN's Jeanne Moos has more now on the Sheen saga and the backlash. JEANNE MOOS, CNN CORRESPONDENT (voice-over): It's the Sheening of America. CHARLIE SHEEN, ACTOR: Welcome to Sheen's Corner. MOOS: He's on every corner. BILL HADER, CAST MEMBER, NBC'S \"SATURDAY NIGHT LIVE\": Live from New York, it's Saturday night. MOOS: Sirius Radio devoted an entire channel to him for a day. ANNOUNCER: Tiger Blood Radio. MOOS: Spike TV will feature Sheen's greatest antics in Taiwanese animation. He's even alienated witches for misusing the word \"warlock.\" UNIDENTIFIED MALE: We bind you. UNIDENTIFIED FEMALE: We bind you. MOOS: So a couple of witches in Massachusetts performed a magical intervention. UNIDENTIFIED FEMALE: We need to come and cleanse your house. MOOS: But Sheen's very own Web casts are what tipped the scale. SHEEN: The tag line is \"torpedoes of truth.\" MOOS (on camera): Well, how's this for a torpedo of truth? It seems the shine has come off Charlie Sheen. (voice-over) In one Web cast he showed off a tattoo on his wrist of his slogan \"winning,\" and said hi to his kids. SHEEN: Daddy loves you. And if you're watching, tell Mom to leave the room. It's on. MOOS: One of his goddesses perched on his lap. Sheen was literally playing with fire as viewers wait for him to combust. SHEEN: It's kind of an eerie image. I'm burning my own face, but I can't feel the MOOS: As one poster on TMZ put it, \"Parents, make your kids watch this. If that doesn't scare them away from drugs, nothing will.\" (on camera) You know the joke has become a little too sick when a comedian refuses to tell any more jokes about Charlie Sheen. (voice-over) Craig Ferguson spoke of how the English insane asylum named Bedlam provided entertainment back in the 18th Century. CRAIG FERGUSON, TALK SHOW HOST: They would pay a penny, and they would look through the peepholes of the cells, and they would look at the lunatics. And looking at the Charlie Sheen thing unfold, and I'm thinking oh, man. MOOS: But Ferguson wasn't kidding. No more Charlie Sheen jokes. Sheen himself has become a verb. The creators of \"South Park\" used it to describe the state they got themselves in when they once dressed in drag for the Oscars. UNIDENTIFIED MALE: We were just Sheening our heads off. MOOS: From our couches, we judge who does the best Sheen. Is it \"SNL\"? HADER: Sorry, middle America. Losers, winning, bye-bye. MOOS: Or Jimmy Fallon? JIMMY FALLON, TALK SHOW HOST: Winning, Adonis DNA. I'm a bitching rock star, blood of a tiger. I'm like Zeus in a Speedo. MOOS: But something stinks when we don't know if it's OK to laugh and winning is a losing proposition. FRED ARMISEN, CAST MEMBER, NBC'S \"SATURDAY NIGHT LIVE\": Winning! HADER: Winning. MILEY CYRUS, SINGER/ACTRESS: Winning. HADER: Winning. MOOS: Jeanne Moos, CNN... SHEEN: Winning, winning, winning! UNIDENTIFIED MALE: Winning, winning, winning! UNIDENTIFIED MALE: Winning, winning, winning! SHEEN: ... New York. BLITZER: Thanks, Jeanne. Thanks very much for watching. I'm Wolf Blitzer in THE SITUATION ROOM. \"JOHN KING USA\" starts right now.
        \\
        \hline
        En Summary & Sheen Fired from Hit Sitcom \\
        \hline
        \multirow{2}{*}{Zh Summary} & \begin{CJK}{UTF8}{gbsn}\small{热门情景喜剧的\textcolor{red}{高光}时刻}\end{CJK} \\
        & (\textcolor{red}{Hightlight} Moment of Hit Sitcom)\\
    \thickhline
     \multicolumn{2}{c}{\textbf{Terminology
}}\\
    \thickhline
     Input Text & Mika: I wanted to ask you to stop supporting the queer group Ann: why? I think they do great things Mika: they discriminated Molly horribly Ann: why? how? Mika: they refused to include her in the panel about sexuality Tom: did they give a reason? Mika: they said her research doesn't match the topic of the panel, what is a bullshit Mika: I believe it's only because she is straight Tom: hmm...\\
     \hline
     {En summary}& The queer group discriminated Molly - they refused to include her in the panel about sexuality. \\
     \hline
     \multirow{2}{*}{Zh summary} & \begin{CJK}{UTF8}{gbsn}\small{\textcolor{red}{同性恋团体}歧视莫莉——他们拒绝让她参加关于性的小组讨论。}\end{CJK}  \\ 
     & (The \textcolor{red}{gay group} discriminated Molly - they refused to include her in the panel about sexuality.)
        \\
     \thickhline
     \multicolumn{2}{c}{\textbf{Coreference}}\\
    \thickhline

    Input Text& Wanda: Let's make a party! Gina: Why? Wanda: beacuse. I want some fun! Gina: ok, what do u need? Wanda: 1st I need too make a list Gina: noted and then? Wanda: well, could u take yours father car and go do groceries with me? Gina: don't know if he'll agree Wanda: I know, but u can ask :) Gina: I'll try but theres no promisess Wanda: I know, u r the best! Gina: When u wanna go Wanda: Friday? Gina: ok, I'll ask"\\
    \hline
     En summary & Wanda wants to throw a party. She asks Gina to borrow her father's car and go do groceries together. They set \textcolor{red}{the date} for Friday. \\
     \hline
     \multirow{2}{*}{Zh summary}&\begin{CJK}{UTF8}{gbsn}\small{旺达想办个聚会。她问吉娜借她父亲的车，两人一起去买聚会用的东西。他们把\textcolor{red}{聚会的日期}定在了周五。}\end{CJK} \\
     & (Wanda wants to throw a party. She asks Gina to borrow her father's car and go do groceries together. They set \textcolor{red}{the date of party} for Friday.) 
        \\
      \thickhline    
    \end{tabular}
    \caption{Error case (a).}
    \label{tab:error1}
\end{table*}

%% file: error2.tex
\begin{table*}[h]
\renewcommand{\arraystretch}{1.5}
    \centering
    \small
    \begin{tabular}{l|p{13.5cm}}
    \thickhline
    \multicolumn{2}{c}{\textbf{Sentence Relation}}\\
    \thickhline
        Input Text  &
        BLITZER: WOLF BLITZER, CNN ANCHOR: Happening now, neck and neck in Indiana. New evidence Barack Obama and Hillary Clinton are in for another fierce battle. Meantime, Obama is dealing with a familiar distraction, the words of his former pastor. We'll tell you what's going on today. John McCain makes a provocative claim about Barack Obama. The Republican suggests Obama is the candidate of the Islamic militant group Hamas. We're looking into this story right now. And President Bush wants to show you the money. We're going to tell you what he's telling taxpayers and why he hopes it will send them to the stores. I'm Wolf Blitzer. You're in THE SITUATION ROOM. Barack Obama wanted to talk to Indiana voters about the soaring gas prices that make their lives tougher every single day, but today the Democrat found he couldn't ignore an ongoing source of controversy. That would be his former pastor, the Reverend Jeremiah Wright. After clamming up and lowering his profile, Wright is now speaking out publicly about the impact on -- and it's having an impact, potentially, at least, on the Obama campaign. Let's go right to CNN's Jessica Yellin. She's watching the story for us. It's a familiar problem over these past few weeks for the senator, Jessica.  JESSICA YELLIN, CNN CONGRESSIONAL CORRESPONDENT: It really has been, Wolf. Today it seems Barack Obama was trying yet again to put that Reverend Wright controversy behind him. He fielded a question about the latest statement from his former pastor.  SEN. BARACK OBAMA (D-IL), PRESIDENTIAL CANDIDATE: I understand that he might not agree with me on my assessment of his comments. That's to be expected. So, you know, he is obviously free to express his opinions on these issues. You know, I've expressed mine very clearly. I think that what he said in several instances were objectionable. And I understand why the American people took offense. And, you know, and as I indicated before, I took offense.  YELLIN (voice-over): Barack Obama speaking out on new comments by his former pastor.  REV. JEREMIAH WRIGHT, BARACK OBAMA'S FMR. PASTOR: And put constantly other and over again...  YELLIN: The Reverend Jeremiah Wright, in an interview airing on PBS Friday night, stands by past sermons that became a political firestorm.  WRIGHT: ... controlled by rich white people.  YELLIN: Wright said his words regarding the 9/11 attacks and race relations were taken out of context. He also reacts to Obama's criticism of him.  WRIGHT: He's a politician. I'm a pastor. We speak to two different audiences. And he says what he has to say as a politician. I say what I have to say as a pastor. Those are two different worlds. I do what ...
        \\
        \hline
        En Summary & Obama's Ex-Pastor Reacts to Criticism; McCain: Obama Favored by Hamas \\
        \hline
        \multirow{2}{*}{Zh Summary} & \begin{CJK}{UTF8}{gbsn}\small{\textcolor{red}{奥巴马前总统回应批评麦凯恩：奥巴马受哈马斯青睐}}\end{CJK} \\
        & (\textcolor{red}{The Ex-President Obama Responds to Criticism of McCain: Obama is Favored by Hamas})\\
    \thickhline
     \multicolumn{2}{c}{\textbf{Others
}}\\
    \thickhline
     Input Text & Elliot: i can't talk rn, i'm rly busy Elliot: can i call u back in about 2 hours? Jordan: Not really, I'm going to a funeral. Jordan: I'll call you tonight, ok? Elliot: sure Elliot: whose funeral is it? Jordan: My colleague's, Brad. Jordan: I told you about him, he had a liver cancer. Elliot: i'm so sorry man, i hope u're ok Elliot: i'll call u at 8 pm \\
     \hline
     {En summary}& Elliot can't talk to Jordan now, he's busy. He'll call him back at 8 pm. Jordan is going to Brad's funeral. He had liver cancer. \\
     \hline
     \multirow{2}{*}{Zh summary} & \begin{CJK}{UTF8}{gbsn}\small{艾略特现在没空和乔丹说话，他很忙。\textcolor{red}{晚上6点}他会给乔丹回电话。乔丹要去参加布拉德的葬礼，布拉德因肝癌去世了。}\end{CJK}  \\ 
     & (Elliot can't talk to Jordan now, he's busy. He'll call him back at \textcolor{red}{8 pm}. Jordan is going to Brad's funeral. He died of liver cancer.)\\
      \thickhline    
    \end{tabular}
    \caption{Error Case (b).}
    \label{tab:error2}
\end{table*}